\documentclass[letterpaper,10pt,journal]{IEEEtran}
\usepackage{blindtext}
\usepackage{graphicx}
\usepackage[bookmarks=true, colorlinks=false,
    linkcolor=black,
    filecolor=magenta,      
    urlcolor=cyan]{hyperref}   
\usepackage{amsmath}
\usepackage{amsfonts}
\usepackage{amssymb}
\usepackage[linesnumbered, ruled,vlined]{algorithm2e}
\SetCommentSty{mycommfont}
\def\Re{{\mathbb{R}}} 
\def\Ne{{\mathbb{N}}}

\author{Opeyemi S.  Orioke$^{1}$, Tauhidul Alam$^{1}$, Joseph Quinn$^{1}$, Ramneek Kaur$^{1}$,\\ Wesam H. Alsabban$^{2}$, Leonardo Bobadilla$^{3}$, and Ryan N. Smith$^{4}$
\thanks{$^{1}$ O. S. Orioke, T. Alam, J. Quinn, and R. Kaur are with the Department of Mathematics, Computer and Information Science, SUNY Old Westbury, NY 11568,  USA. (e-mail: \{oorioke, alamt, jquinn14,    rkau23\}@oldwestbury.edu)
}

\thanks{$^{2}$
W. H. Alsabban is with the Computer and Information Systems, Umm Al-Qura University, Makkah, Saudi Arabia. (e-mail: whsabban@uqu.edu.sa)
}
\thanks{$^{3}$
 L. Bobadilla is with the School of Computing and Information Sciences, Florida International University, Miami, FL 33199, USA. (e-mail: bobadilla@cs.fiu.edu)
}
\thanks{$^{4}$ R. N. Smith is with the Department of Physics and Engineering, Fort Lewis College, Durango, CO 81301, USA. (e-mail: rnsmith@fortlewis.edu)}}
\begin{document}
%
\title{\vspace{8pt}Feedback Motion Planning for Long-Range Autonomous Underwater Vehicles}

\maketitle

\begin{abstract}
Ocean ecosystems have spatiotemporal variability and dynamic complexity that require a long-term deployment of an autonomous underwater vehicle for data collection. A new long-range autonomous underwater vehicle called Tethys is adapted to study different oceanic phenomena. 
Additionally, an ocean environment has external forces and moments along with changing water currents which are generally not considered in a vehicle kinematic model. In this scenario, it is not enough to generate a simple trajectory from an initial location to a goal location in an uncertain ocean as the vehicle can deviate from its intended trajectory. As such,  we propose to compute a feedback plan that adapts the vehicle trajectory in the presence of any modeled or unmodeled uncertainties. In this work, we present a feedback motion planning method for the Tethys vehicle by combining a predictive ocean model and its kinematic modeling. Given a goal location, the Tethys kinematic model, and the water flow pattern, our method computes a feedback plan for the vehicle 
in a dynamic ocean environment that reduces its energy consumption. The computed feedback plan provides the optimal action for the Tethys vehicle to take from any location of the environment to reach the goal location considering its orientation. Our results based on actual ocean model prediction data demonstrate the applicability of our method.

\end{abstract}
\maketitle

\section{Introduction}

Ocean ecosystems are complex and have high variability in both time and space. Consequently, ocean scientists must collect data over long time periods to obtain a synoptic view of ocean ecosystems and resolve their spatiotemporal variability. On the other hand, autonomous underwater vehicles (AUVs) are increasingly being used for studying different phenomena in ocean ecosystems such  as oil spill mapping~\cite{kinsey2011assessing}, harmful algal blooms~\cite{das2010towards}, phytoplankton and zooplankton communities~\cite{kalmbach2017phytoplankton}, and coral bleaching~\cite{manderson2017robotic}. These AUVs are classified into two categories: (i) propeller-driven vehicles such as Dorado that can move fast and gather numerous sensor observations; and (ii)  minimally-actuated vehicles such as Drifters, Profiling floats, and Gliders that can remain on deployment for long time periods, ranging from many days to many weeks. 
\begin{figure}
  \begin{center}
  \hspace{-3pt}\includegraphics[scale=0.126]{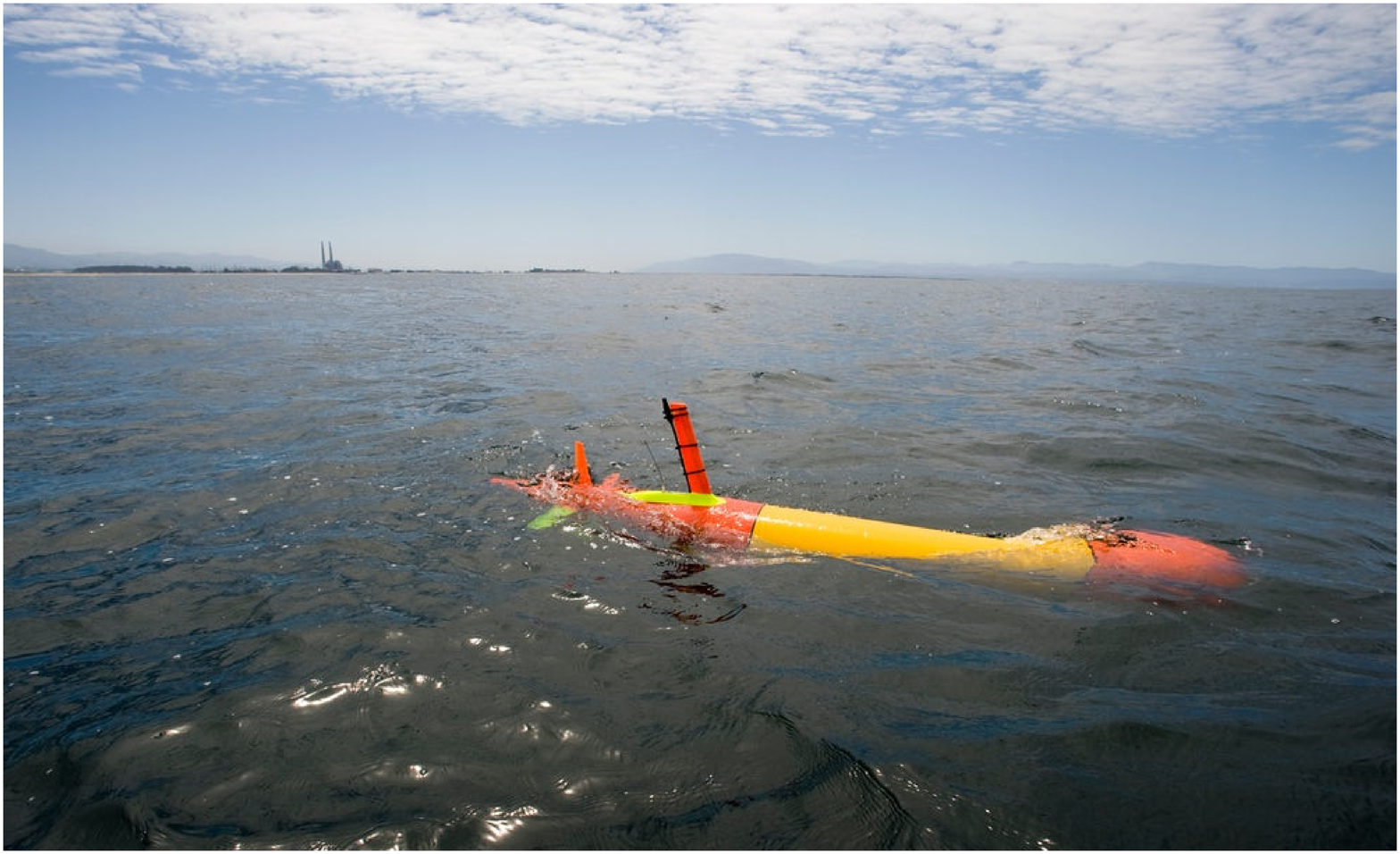}
    \includegraphics[scale=0.48]{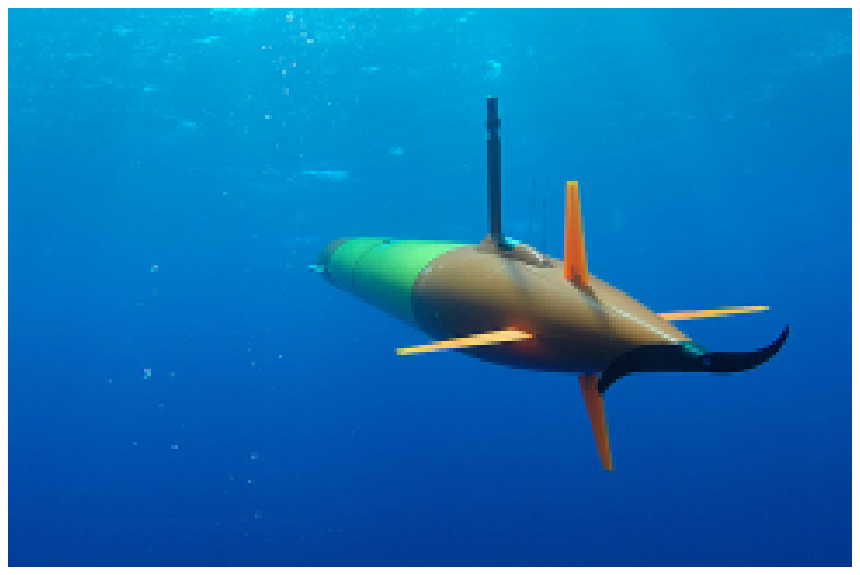}
  \end{center}
\caption{\label{fig:tethys} Two Tethys vehicles observing the oceanic environments~\cite{auv,lrauv}.}
\vspace{2pt}
\end{figure}

Therefore, one option to perform persistent ocean observation is to utilize minimally-actuated underwater vehicles. However, such vehicles do not have the capabilities to control their actuation, and they move slowly with  water currents. Furthermore, these minimally-actuated vehicles cannot navigate through water currents properly while executing a prescribed path or mission due to their limited actuation capabilities. Another option for ocean monitoring is to use propeller-driven AUVs. Nonetheless, their mission endurance is limited to only a few days, and they consume a lot of energy for actuation.

 A new breed of the long-range autonomous underwater vehicle (LRAUV) called Tethys combines the advantages of both minimally-actuated and propeller-driven AUVs~\cite{hobson2012tethys}. The Tethys AUV can move quickly for hundreds of kilometers, float with water currents, and carry a broad range of sensors for data collection. It can also control its buoyancy for changing depths in the water and the angle at which it moves through the water. Most importantly, this vehicle can be deployed in the water for weeks at a time. Two examples of deployed Tethys AUVs to observe oceans are shown in Fig.~\ref{fig:tethys}. A planning and control technique for this vehicle is  critical to increase its autonomy and  generate mission trajectories during long-range operations. The execution of a planned trajectory for this vehicle is also challenging due to the variability and uncertainty of ocean currents. Thus, it is  not enough to generate a simple navigation trajectory from an initial location to a goal location in a dynamic ocean environment with unmodeled uncertainties as the vehicle can deviate from its trajectory. In other words, this navigation trajectory suffers from simulation bias as there is no sensor feedback during the trajectory implementation.

 
As such, we consider the use of feedback motion planning for the Tethys vehicle by combining a predictive ocean model and its kinematic modeling. It is also not possible to model all the uncertainties of a dynamic ocean such as gravity, buoyancy, hydrodynamic and other uncertain
forces and moments. 
Hence, the motivation for our feedback motion planning based method is to compute a {\em feedback plan}. A feedback plan is computed over the entire state space of a vehicle, so the vehicle can adapt its trajectory from any deviated state in the presence of any noise or modeling errors. Furthermore, this feedback plan is crucial when the initial state of an AUV in a trajectory is unknown as the plan indicates what action to take from every state.



{\em Statement of contributions:} In this work, first, we develop a kinematic model of the Tethys and generate flow fields from ocean model predictions. Then, we present a method to compute a feedback plan from the Tethys kinematic model and the water flow pattern that drives the vehicle from any location to the goal location in a dynamic ocean environment and  reduces its energy consumption.

\section{Related Work}
\label{sec:rw}
The feedback mission control of autonomous underwater vehicles in dynamic and spatiotemporal underwater environments has attracted a great deal of interest. A feedback trajectory tracking
scheme was developed for an AUV in a dynamic environment like an ocean
with the presence of several modeled and unmodeled uncertainties~\cite{sanyal2009robust}. In~\cite{reis2018feedback}, an informative feedback plan was generated for AUVs to visit essential locations by estimating Kriging Errors from spatiotemporal fields. In this method, a feedback plan was calculated via
the Markov Decision Model (MDP) based problem formulation.  A finite state automata based supervisory feedback control was presented  in~\cite{xu2009auv} for obstacle avoidance by an AUV.    A
temporal plan was calculated in ~\cite{cashmore2014auv} for AUV mission control that optimizes the time taken to complete a single inspection tour.  
In~\cite{cashmore2014auv}, a feedback and replanning framework was integrated along with the temporal plan in  the  Robot  Operating  System  (ROS). In~\cite{caldwell2010motion}, Sampling
Based Model Predictive Control (SBMPC)   is utilized to simultaneously generate control inputs and  feasible trajectories  for  an AUV in the presence of a number of nonlinear constraints.

Open-loop trajectory design methods~\cite{smith2010planning,chyba2009increasing}  drive a AUV from a given initial location to the desired goal location minimizing a cost in terms of energy and time taken by the vehicle. 
The implementation of open-loop trajectories for AUVs work well in  environments without any model uncertainties. In our previous work~\cite{alam2018deployment}, we proposed an open-loop approach for solving the problem of deploying a set of minimally-actuated drifters for persistent monitoring of an aquatic environment. In our recent work~\cite{alam2018underactuated}, we predicted the localized trajectory of a drifter for a sequence of compass observations during its deployment in a marine environment. We presented a closed-loop approach in~\cite{alam2018underactuated} when an AUV has a considerable unpredictability of executing its action in a dynamic marine environment. Moreover, the previous studies~\cite{hobson2012tethys, bellingham2010efficient} on the Tethys AUV described the mission and other capabilities of the vehicle. However, there is no work on the development of a planning algorithm for controlling the vehicle.  In this work, we utilize the Tethys vehicle's sensor feedback to control its mission operation taking into account its several drifting and actuation capabilities as well as minimizing its energy consumption. 


\section{Model and Problem Definition}
\label{sec:pre}

In this section, we present the kinematic model for the Tethys vehicle in a representation of an environment and formulate our problem of interest.

\subsection{Model Definition}
We consider a $3$-D  environment where a workspace is an ocean environment denoted as ${\cal W} \subset \Re^3$. 
The workspace is divided into a set of $2$-D water current layers at different depths  of the environment which represents the third dimension. Let $L$ be the total number of water current layers in the environment. Hence, the workspace can be defined as ${\cal W}= W_1 \cup W_2 \cup \cdots  \cup W_L$.
At each current layer, we model the workspace $W_l \subset \Re^2$, where $l\in\{1,\ldots,L\}$, as a polygonal environment. Let $O_l \subset \Re^2$ be the land and littoral region of the environment at each layer which is considered an inaccessible region for the vehicle. The free-water space at each current layer is composed of all navigable locations for the vehicle, and it is defined as $E_l =  W_l \setminus O_l$. 
The free water space in the whole workspace is denoted by ${\cal E}=E_1 \cup E_2 \cup \cdots  \cup E_L$.

We discretize each workspace layer $W_l$ as a $2$-D grid. Each grid point or location, denoted as $q$, has a geographic coordinate in the form of longitude, latitude, and depth $q = (x_t,y_t,l_t)$, where $x_t,y_t \in \Re$ and  $l_t \in \{1,\ldots,L\}$.  
The geographic coordinate of each $q$ represents the center of an equal-sized grid location. The state space for the vehicle is defined as $X = {\cal E} \times S^{1}$ in which $S^{1}$ is the set of angles in [$0, 2\pi$) that
represents the vehicle's orientations. A state of the vehicle in the state space is indexed by a cell index $z \in \{1, \ldots, N\}$, where
$N$ represents the total number of cells in $X$.
A cell in the state space is represented as $z = (q, \theta)$ in which $q$ denotes the vehicle position, and $\theta$ provides the vehicle orientation. 
Let $Z = \{1, \ldots, N\}$ denote the set of all cells in the state space. Let
$x_G \in X$ be a goal state or location of the vehicle at any water current layer of the environment.

\subsection{Problem Formulation}
\begin{figure}

  \begin{center}
    \hspace{-5pt}\includegraphics[scale=.48]{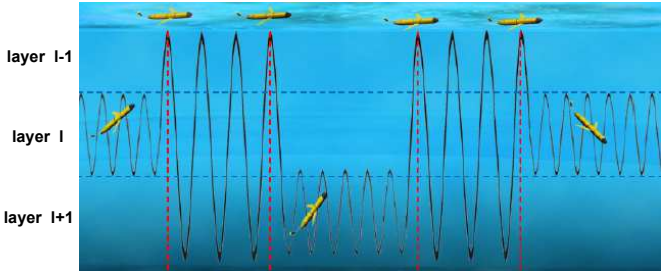}
  \end{center}

\caption{\label{fig:glide} A glider is moving up and down by transitioning between two water current layers~\cite{al2012extending}.}

\end{figure}

We consider the unicycle motion model for the Tethys vehicle. In our vehicle motion model, we assume that the Tethys can drift taking advantage of water currents, actuate to move forward with or against currents, rotate left and right in the same location, glide up and down by transitioning between two water current layers  as
illustrated in Fig.~\ref{fig:glide}. We also incorporate noise and uncertainty in the vehicle's movement to account for the modeling error and unmodeled dynamics. Hence, the {\em action set}
of the vehicle in the underwater environment is $U = \{\text{drift, move forward, rotate left, rotate right,  up, 
 down} \}$. Each action $u \in U$ has an energy cost associated with it which is denoted as $c_u$. In particular, we consider  $c_{\text{drift}}< c_{\text{glide}}< c_{\text{forward}} < c_{\text{rotate}}$ for different actions of the Tethys vehicle. Let the {\em cost set} for consuming energy in executing different actions by the Tethys vehicle be $C$.
A  feedback plan $\pi$ is defined as a function $\pi : X \rightarrow U $ which produces an action $u = \pi(x) \in U(x)$, for any state $x \in  X$, to reach the goal state $x_G$. When the vehicle does not know its initial state, it is appropriate to find a feedback plan that maps every state to an action. A feedback plan is called a {\em solution} to the problem if it causes the goal state to be reached from every state in $X$.  This motivates us to formulate the following problem.

{\bf Problem 1. Computing a feedback plan for an LRAUV: }
{\em Given an ocean environment ${\cal E}$,  the action set of the Tethys vehicle $U$, the vehicle motion model, the water flow pattern,  and a goal location $x_G$, compute a feedback plan $\pi$ that drives the vehicle for any location of the environment  to reach the goal location $x_G$ minimizing the energy cost.
}

\section{Method}
\label{sec:meth}
In this section, we detail our method for solving the problem formulated in Section~\ref{sec:pre}.

\subsection{Data Acquisition}
\begin{figure} [ht!]
  \begin{center}
     \includegraphics[width=0.4\textwidth]{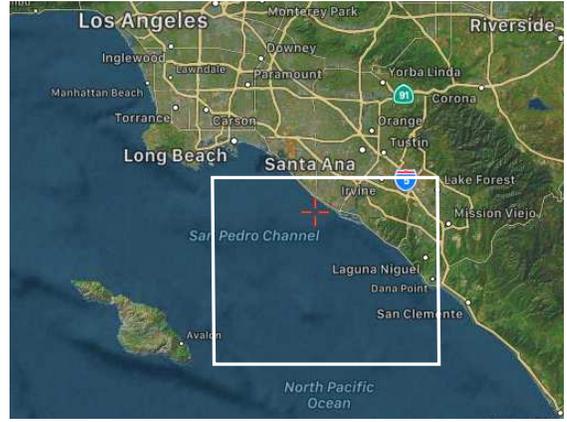}
  \end{center} 
 \caption{\label{fig:area} The area of interest in the SCB region, California.}
 \end{figure}

We use the Regional Ocean Modeling System (ROMS)~\cite{shchepetkin2005regional} predicted oceanic current data 
in the Southern California Bight (SCB) region, CA, USA,  as illustrated in Fig.~\ref{fig:area}, which is contained within $33^\circ 17' 60''$ N to $33^\circ 42'$ N and $-117^\circ 42'$ E to $-118^\circ 15' 36''$ E. ROMS is a free-surface, split-explicit, terrain-following, nested-grid mode, and  an extensively used ocean model. ROMS is also an open-source ocean model that is widely accepted and supported throughout the oceanographic and modeling communities. Furthermore, the model was developed to study ocean processes along the western U.S. coast which is our area of interest.  
The four dimensions of $4$-D ROMS current velocity prediction data are longitude, latitude, depth, and time. The ROMS current velocity prediction data are given at depths from $0$ m to $125$ m and with $24$ hours forecast for each day.  The three velocity components of oceanic currents are northing current ($\alpha$), easting current ($\beta$), and vertical current ($\gamma$).  These velocity components are given based on the four dimensions (time, depth, longitude, and latitude). We use  velocity prediction data of water currents at depths between $0$ m and $15$ m for a specific time.

\subsection{Generating the Water Flow Pattern}

 We create flow fields at several water current layers from the ROMS ocean current prediction data.  Ocean current velocity prediction data for a specific time and at a particular water current layer can be represented as a flow field. 
Let the flow field on a location $q$ at a particular water current layer of the environment $ E_l$  be $F(q)$.  
For a location $q$ at a  particular water current layer, the easting velocity component along the latitude axis is denoted by $\alpha(q)$, the northing velocity component along the longitude axis is denoted by $\beta(q)$, and the vertical velocity component at that water current layer is denoted by $\gamma(q)$. 
The flow field based on three velocity components for a location $q$ at that water current layer  is specified as: 
\begin{equation}
F(q) = [\alpha(q), \beta(q), \gamma(q)]. 
\end{equation}
The vertical velocity component of the ocean current $\gamma(q)$ at all water current layers is considered zero. Thus, we create flow fields for all water current layers. Then, we find flow lines of the water flow from these flow fields.
Flow lines of the water flow over the flow field $F$ are the trajectories or paths traveled by an omnidirectional vehicle at the given water current layer whose velocity field is the flow field.

To find the water flow pattern, we need all flow lines from locations at different water current layers for a small time step $\Delta t$  so that we can map one location to another based on these flow lines.  To calculate the next mapped location $z(\Delta t)$ after a small time interval $\Delta t$ from each initial location at time zero $z(0)$, we use the Euler integration method as follows:

\begin{equation}
q(\Delta t) = q(0)+ \Delta t \hspace{2pt} F(q(0)). 
\end{equation}

It gives the endpoint of the flow line from the initial location $q$ after the small time $\Delta t$. After that, we use the Euclidean distance for locating the nearest location from this endpoint. This nearest location $q'$ becomes the next mapped location of the initial location $q$. Following this process, we obtain all flow lines for the small time step $\Delta t$ at each water current layer. Finally, we get the next mapped location for each location in the environment ${\cal E}$ at all water current layers.



\begin{algorithm}
\caption{\label {alg:feedbackplan}\textsc{ComputeFeedbackPlan}   (${\cal E},F,U, C, x_G$)}
\DontPrintSemicolon
\KwIn{${\cal E},F,U,C,x_G$ -- Environment, flow fields, action set, cost set, goal  state}
\KwOut{$\pi$ -- A feedback plan}
$G.V \gets \emptyset, \quad G.E \gets \emptyset, \quad \pi \gets \emptyset$\;
\For{$i \gets 1 \text{ to } N$}{
$z\gets i$ \;
$q,\theta \gets \textsc{CellState}(z)$\;

$q' \gets \textsc{MappedCell}(q, F)$\;
$\theta_w \gets \textsc{WaterFlowDirection}(q,q')$\;
$s \gets \textsc{AlignmentScore}(\theta,\theta_w)$\;

\For{$j \gets 1 \text{ to } |U|$}{
$Z' \gets \textsc{MappedCells}(s,\theta,q)$ \tcp*[f]{Add uncertainty}\;
$G.V\gets G.V \cup Z' $\;
$G.E\gets G.E \cup \{(z,z')\mid z'\in Z'\}$\;
$\forall z' \in Z': w(z, z') \leftarrow c_j$ \;
}
}
\For{$k \gets 1 \text{ to } N$}{
$\tau \leftarrow \textsc{ShortestPath}(G, x_k, x_G)$\;
$\pi(x_k) \leftarrow \textsc{ComputeAction}(\tau)$\;

}

\Return{$\pi$  }\;

\end{algorithm}

\subsection{Computing a Feedback Plan}

In this step, we compute a feedback plan  for an LRAUV. To obtain the feedback plan,   Algorithm~\ref{alg:feedbackplan} takes as input the environment ${\cal E}$, flow fields $F$, the action set $U$, the cost set $C$, and a goal state $x_G$. It returns a feedback plan $\pi$. Let the water flow direction be $\theta_w$. We define the vehicle's alignment score $s$ with respect to the water flow direction. Let the shortest path for the vehicle be $\tau : [0, 1] \rightarrow X$ such that $\tau(0) = x_I$, $\tau(1) = x_G$. A directed graph is defined as $G=(V,E)$, where each vertex $v \in V$ represents a state $x$ of the vehicle and   an edge $(v,v')\in E$, where $v, v' \in V$, represents a state transition from one state $x\in X$ to another state $x'\in X$. We define the weight function $w : E 
\rightarrow \Ne^+$ that assigns the non-negative edge weight.

From each cell $z \in Z$, Algorithm~\ref{alg:feedbackplan} finds the geographic location $q$ and the vehicle orientation $\theta$  (line $4$).  From this geographic location $q$, it obtains the next prospective location $q'$ using the flow fields $F$ (line $5$).
 Next, we calculate the water flow direction $\theta_w$ from the previous location $q$ and the new location $q'$ (line $6$). The angular distance $\rho$ between $\theta$ and $\theta_w$ is calculated as: 
 \begin{equation}
\rho(\theta, \theta_w) = \text{min} (|\theta - \theta_w|,  2\pi - |\theta - \theta_w|), 
\end{equation}
for which $\theta, \theta_w \in [0, 2\pi]$. As the movement of the vehicle will depend on the vehicle's alignment with respect to the water flow direction,  we calculate the alignment score $s$ between the vehicle orientation $\theta$ and the water flow direction $\theta_w$ for each cell $z$ (line $7$) as follows:
  \begin{equation}
 s = \rho(\theta, \theta_w)/180.
 \end{equation}
 The alignment score $s$ for the vehicle ranges from $0$ to $1$. Considering the vehicle's uncertain motion, alignment score $s$, the vehicle orientation $\theta$, and the initial location $q$, it computes the set of next mapped cells $Z'$, where $Z' \subset Z$, for each action $u \in U $. In a boundary location of the environment, we include the drifting action for the vehicle to move its position to one of the neighboring locations.
The set of next mapped cells $Z'$ are added to the vertices  and their ordered pairs $(z,z')$, where $z'\in Z'$, are included to the edges of the graph $G$. Each edge $(v,v')\in E$ represents a state transition to reach a vertex $v'$ starting from a vertex $v$ applying one action
$u \in U$.  
The edge weights are assigned according to the energy required for the actions of the vehicle (lines $8$ -- $12$).    

Once we have the graph $G$, we apply a modified version of the Dijkstra's algorithm to find the shortest path $\tau$ from a given initial location to the goal location. The first state transition in the path $\tau$ will provide the best action for the vehicle to take at the given initial location. Repeating the same process for all the initial locations, we eventually compute a feedback plan $\pi$ 
that provides the optimal action for the Tethys vehicle to take from any location of the environment to reach the goal location $x_G$ (lines $13$ -- $15$). 

\section{Results}
\begin{figure}[ht!]
\begin{center}
\begin{tabular}{c}
\hspace{-5pt}\includegraphics[scale=0.45]{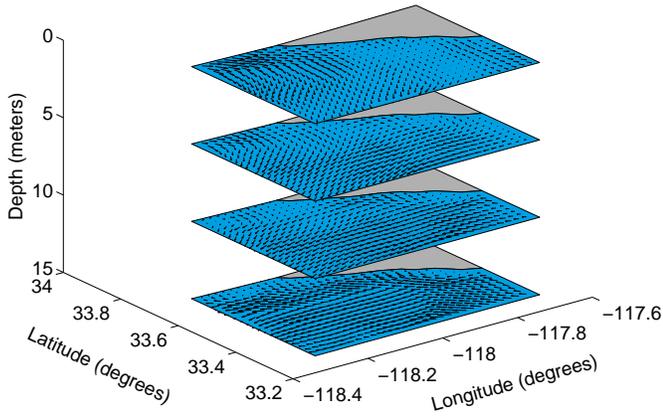}\\
\end{tabular}
\end{center}
\caption{\label{fig:results1vec} Flow fields generated from ROMS  oceanic current prediction data~\cite{shchepetkin2005regional}.  }

\end{figure}

\begin{figure}[ht!]
\begin{center}
\begin{tabular}{c}
\hspace{-5pt}\includegraphics[scale=0.45]{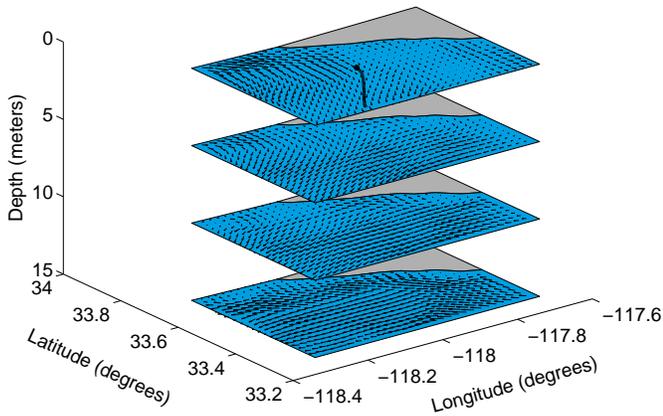}\\
\end{tabular}
\end{center}
\caption{\label{fig:results1stream} An example flow line generated from  flow fields. }

\end{figure}

We tested our method through simulations using the ROMS~\cite{shchepetkin2005regional} predicted ocean current data observed in the SCB region. The $3$-D ocean environment was taken into account as the simulation environment for the Tethys movements having three $2$-D ocean surfaces at four different water current layers or depths (e.g., $0$ m, $5$ m, $10$ m, $15$ m).  Each  $2$-D ocean surface is tessellated into a grid map. The resolution of the grid map is $21\times29$. Four flow fields from the ocean current data  are calculated at four water current layers of the environment as shown in Fig.~\ref{fig:results1vec}. The flow lines of the ocean current data are generated through the Euler numerical integration method from all locations of the water surfaces for a small time $\Delta t$ which is the time to pass a cell. A flow line  for a calculated flow field from a particular location  for all time $t$ is illustrated in
Fig.~\ref{fig:results1stream}.

\begin{figure}[ht!]

\begin{center}
\begin{tabular}{c}

\hspace{-5pt}\includegraphics[scale=0.45]{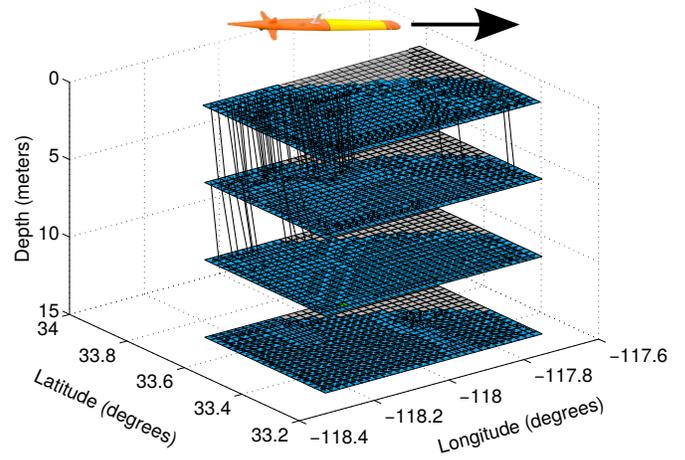} \\ 

\end{tabular}
\end{center}

\caption{\label{fig:results1policy} 
A computed feedback plan showing the best actions with black arrows from all locations of the environment while the vehicle is oriented eastward at its initial deployment. }
\end{figure}

\begin{figure}[ht!]

\begin{center}
\begin{tabular}{c}

\hspace{-5pt}\includegraphics[scale=0.45]{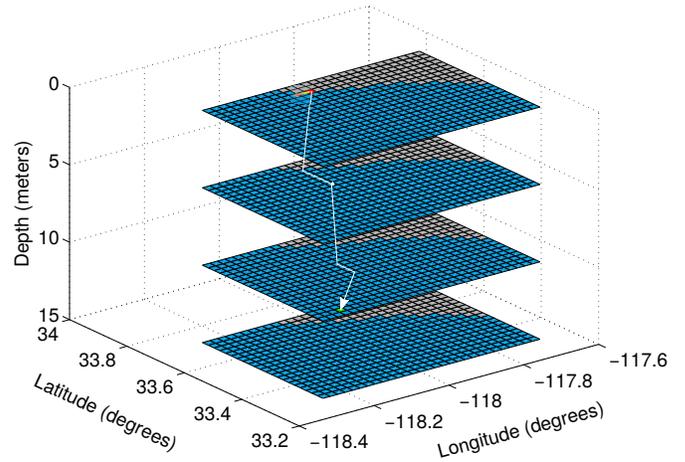} \\

\end{tabular}
\end{center}

\caption{\label{fig:results1navpath} A calculated navigation trajectory delineated with the white line from the red initial location to the green goal location applying the feedback plan. }
\vspace*{-15pt}

\end{figure}

We implemented Algorithm~\ref{alg:feedbackplan} to compute the feedback plan for the Tethys vehicle considering its orientation. In our implementation, we considered two grid tiles forward movement along the orientation of the vehicle and two grid tiles drifting or gliding movement along the water flow direction when the alignment score is $s \leq 0.2$. In the case of the alignment score $0.2 < s \leq 0.5$, we considered one grid tile forward movement along the orientation of the vehicle and one grid tile drifting or gliding movement along the water flow direction. In the case of the alignment score $s > 0.5$, we considered in our implementation that the vehicle does not change its position rather floats at the current position. We also implemented the in-place rotation of the vehicle by changing the vehicle  orientation $45^{\circ}$ to the left (counterclockwise rotation) or $45^{\circ}$  to the right (clockwise rotation).
The costs set in our implementation for different actions of the vehicle are $c_{\text{drift}}=0$, $c_{\text{glide}}=2$,  $c_{\text{forward}}=4$, and $c_{\text{rotate}}=10$. A computed feedback plan is shown in Fig.~\ref{fig:results1policy} that illustrates the best actions from all locations of the environment when the vehicle's orientation is eastward at its initial deployment.
A calculated navigation trajectory between an initial location and a goal location in the environment applying the computed feedback plan is illustrated in Fig.~\ref{fig:results1navpath}.


\section{Conclusion and Future Direction}
\label{sec:conc}
In this paper, we proposed a feedback motion planning method by combining predicted water current data and the Tethys vehicle kinematic model. 
The computed feedback plan
adapts the vehicle trajectory if necessary in an ocean environment
with several uncertainties and reduces the vehicle energy
consumption as the plan utilizes the vehicle's drifting and
other actuation capabilities. First, we generated the water flow
pattern from water current prediction data. Second, a directed
graph was created based on the water flow pattern and
the vehicle kinematic model and we applied the shortest path
algorithm to find paths from all initial locations to a goal
location. Finally, a feedback plan was computed that drives
the vehicle from any location to the goal location of the
environment using the best possible actions.

In the future, we can analyze the temporal variability of
our computed feedback plan. We plan to evaluate our method
with a quantitative study incorporating several uncertainties
and noises in executing a navigation trajectory in our simulated
environment.

\section*{Acknowledgments}
This work was supported by the National Science Foundation via NSF MRI 1531322, the US Office of Naval Research Award N000141612634, and the Deanship of Scientific Research at Umm Al-Qura University Award 3-2-0002-COM-15. This work is also supported in part by the U.S. Department of Homeland Security under Grant Award Number 2017-ST-062000002.

\bibliographystyle{ieeetr} 
{
\bibliography{oceans} 

\begin{thebibliography}{10}

\bibitem{kinsey2011assessing}
J.~C. Kinsey, D.~R. Yoerger, M.~V. Jakuba, R.~Camilli, C.~R. Fisher, and C.~R.
  German, ``Assessing the deepwater horizon oil spill with the sentry
  autonomous underwater vehicle,'' in {\em Proceedings of the IEEE/RSJ
  International Conference on Intelligent Robots and Systems (IROS)},
  pp.~261--267, 2011.

\bibitem{das2010towards}
J.~Das, K.~Rajany, S.~Frolovy, F.~Pyy, J.~Ryany, D.~A. Caronz, and G.~S.
  Sukhatme, ``Towards marine bloom trajectory prediction for {AUV} mission
  planning,'' in {\em Proceedings of the IEEE International Conference on
  Robotics and Automation (ICRA)}, pp.~4784--4790, 2010.

\bibitem{kalmbach2017phytoplankton}
A.~Kalmbach, Y.~Girdhar, H.~M. Sosik, and G.~Dudek, ``Phytoplankton hotspot
  prediction with an unsupervised spatial community model,'' in {\em
  Proceedings of the IEEE International Conference on Robotics and Automation
  (ICRA)}, pp.~4906--4913, 2017.

\bibitem{manderson2017robotic}
T.~Manderson, J.~Li, N.~Dudek, D.~Meger, and G.~Dudek, ``Robotic coral reef
  health assessment using automated image analysis,'' {\em Journal of Field
  Robotics}, vol.~34, no.~1, pp.~170--187, 2017.

\bibitem{auv}
Autonomous underwater vehicles. Available at
  \href{https://www.mbari.org/at-sea/vehicles/autonomous-underwater-vehicles/}{https://www.mbari.org/at-sea/vehicles/autonomous-underwater-vehicles/}.

\bibitem{lrauv}
Self-driving robots collect water samples to create snapshots of ocean
  microbes. Available at
  \href{https://www.mbari.org/at-sea/vehicles/autonomous-underwater-vehicles/}{https://www.mbari.org/at-sea/vehicles/autonomous-underwater-vehicles/}.

\bibitem{hobson2012tethys}
B.~W. Hobson, J.~G. Bellingham, B.~Kieft, R.~McEwen, M.~Godin, and Y.~Zhang,
  ``Tethys-class long range {AUV}s-extending the endurance of propeller-driven
  cruising {AUV}s from days to weeks,'' in {\em Proceedings of the IEEE/OES
  Autonomous Underwater Vehicles Symposium}, pp.~1--8, 2012.

\bibitem{sanyal2009robust}
A.~K. Sanyal and M.~Chyba, ``Robust feedback tracking of autonomous underwater
  vehicles with disturbance rejection,'' in {\em Proceedings of the American
  Control Conference}, pp.~3585--3590, 2009.

\bibitem{reis2018feedback}
G.~M. Reis, T.~Alam, L.~Bobadilla, and R.~N. Smith, ``Feedback-based
  informative {AUV} planning from {K}riging errors,'' in {\em Proceedings of
  the IEEE/OES Autonomous Underwater Vehicles}, 2018.

\bibitem{xu2009auv}
H.~Xu and X.~Feng, ``An {AUV} fuzzy obstacle avoidance method under event
  feedback supervision,'' in {\em Proceedings of the MTS/IEEE OCEANS},
  pp.~1--6, 2009.

\bibitem{cashmore2014auv}
M.~Cashmore, M.~Fox, T.~Larkworthy, D.~Long, and D.~Magazzeni, ``{AUV} mission
  control via temporal planning,'' in {\em Proceedings of the IEEE
  International Conference on Robotics and Automation (ICRA)}, pp.~6535--6541,
  2014.

\bibitem{caldwell2010motion}
C.~V. Caldwell, D.~D. Dunlap, and E.~G. Collins, ``Motion planning for an
  autonomous underwater vehicle via sampling based model predictive control,''
  in {\em Proceedings of the MTS/IEEE OCEANS Seattle}, pp.~1--6, 2010.

\bibitem{smith2010planning}
R.~N. Smith, Y.~Chao, P.~P. Li, D.~A. Caron, B.~H. Jones, and G.~S. Sukhatme,
  ``Planning and implementing trajectories for autonomous underwater vehicles
  to track evolving ocean processes based on predictions from a regional ocean
  model,'' {\em The International Journal of Robotics Research}, vol.~29,
  no.~12, pp.~1475--1497, 2010.

\bibitem{chyba2009increasing}
M.~Chyba, T.~Haberkorn, S.~Singh, R.~Smith, and S.~Choi, ``Increasing
  underwater vehicle autonomy by reducing energy consumption,'' {\em Ocean
  Engineering}, vol.~36, no.~1, pp.~62--73, 2009.

\bibitem{alam2018deployment}
T.~Alam, G.~M. Reis, L.~Bobadilla, and R.~N. Smith, ``A data-driven deployment
  approach for persistent monitoring in aquatic environments,'' in {\em
  Proceedings of the IEEE International Conference on Robotic Computing (IRC)},
  pp.~147--154, 2018.

\bibitem{alam2018underactuated}
T.~Alam, G.~M. Reis, L.~Bobadilla, and R.~N. Smith, ``An underactuated vehicle
  localization method in marine environments,'' in {\em Proceedings of the
  MTS/IEEE OCEANS Charleston}, pp.~1--8, 2018.

\bibitem{bellingham2010efficient}
J.~G. Bellingham, Y.~Zhang, J.~E. Kerwin, J.~Erikson, B.~Hobson, B.~Kieft,
  M.~Godin, R.~McEwen, T.~Hoover, J.~Paul, {\em et~al.}, ``Efficient propulsion
  for the {T}ethys long-range autonomous underwater vehicle,'' in {\em
  Proceedings of the IEEE/OES Autonomous Underwater Vehicles}, pp.~1--7, 2010.

\bibitem{al2012extending}
W.~H. Al-Sabban, L.~F. Gonzalez, and R.~N. Smith, ``Extending persistent
  monitoring by combining ocean models and {M}arkov decision processes,'' in
  {\em Proceedings of the MTS/IEEE OCEANS}, pp.~1--10, 2012.

\bibitem{shchepetkin2005regional}
A.~F. Shchepetkin and J.~C. McWilliams, ``The {R}egional {O}ceanic {M}odeling
  {S}ystem (\uppercase{ROMS}): a split-explicit, free-surface,
  topography-following-coordinate oceanic model,'' {\em Ocean Modelling},
  vol.~9, no.~4, pp.~347--404, 2005.

\end{thebibliography}
}

\end{document}